\RequirePackage{amsmath}
\documentclass[runningheads,a4paper]{lncs/llncs}
\usepackage{multibib}
\usepackage{graphicx}
\usepackage{tabularx}
\usepackage{soul}
\usepackage{times}
\usepackage{url}
\usepackage{latexsym}
\usepackage{layout}
\usepackage{float}
\usepackage{placeins}
\usepackage{wrapfig}
\usepackage{afterpage}
\usepackage[normalem]{ulem}
\usepackage{color}
\usepackage{nicefrac}
\usepackage{multirow}
\usepackage{amsmath}
\usepackage{pgfplots}
\pgfplotsset{compat=1.14}
\usepackage{tikzscale}
\usepackage[utf8]{inputenc}
\usepackage[T2A]{fontenc}
\usepackage[russian,USenglish]{babel}

\usepackage{epstopdf}

\usepackage{hyperref}
\usepackage{xstring}
\usepackage{ifdraft}

\begin{document}
\mainmatter  % start of an individual contribution
\title{Debugging Neural Machine Translations}
\titlerunning{Debugging Neural Machine Translations}

\author{Matīss Rikters}
\authorrunning{M. Rikters}
% \author{Author names are omitted for blind review}

\institute{Tilde \\
          Vien\={\i}bas gatve 75A, Riga, Latvia, LV-1004 \\
         \email{\{matiss.rikters\}@tilde.lv}\\}
\maketitle

\begin{abstract}
In this paper, we describe a tool for debugging the output and attention weights of neural machine translation (NMT) systems and for improved estimations of confidence about the output based on the attention. The purpose of the tool is to help researchers and developers find weak and faulty example translations that their NMT systems produce without the need for reference translations. Our tool also includes an option to directly compare translation outputs from two different NMT engines or experiments. In addition, we present a demo website of our tool with examples of good and bad translations: \url{http://attention.lielakeda.lv}.
\keywords{Neural machine translation, Visualization tool, Attention mechanism}
\end{abstract}

\section{Introduction}
\label{intro}

As one of the primary use-cases for the modern computer - automated translation of texts from one language into another or machine translation (MT) has evolved vastly since its early days in the 1950s. There have been several large paradigm shifts that have greatly impacted the field of MT - rule-based MT (RBMT), statistical MT (SMT) and neural network MT (NMT) \cite{nmt}. With each paradigm shift detailed understanding of how the system produces its final translation has changed from fully clear in the case of RBMT to slightly less, but often still predictable in SMT, to often completely unpredictable in NMT. Many of the existing tools for inspecting results of statistical phrase-based approaches are either not compatible or serve little purpose in dealing with neural network generated output.

In this paper, we propose a tool for browsing, inspecting and comparing translations  specifically designed for NMT output. The tool uses the attention weights that correspond to
specific token pairs which are generated during the decoding process, by turning them into one of several visual representations that can help humans better understand how the
output translations were produced.
Aside from just visualizing attention alignments, the tool also uses them to estimate the confidence in translation, which allows to distinguish acceptable outputs from completely unreliable ones. For this no reference translations are required.

The structure of this paper is as follows: Section \ref{sec:related} summarizes related work on tools for inspecting translation outputs and alignments; Section \ref{sec:tool} introduces some concepts of the baseline tool - how it scores translations and displays the visualizations in different environments, as well as outlines the improvements made to make it more useful for debugging machine translation output. In section \ref{sec:debugging} we give an overview of how to make the most use of our tool in finding odd translations, what to look for when comparing them and possible causes of errors. Finally, we conclude in Section \ref{sec:conclusion} and introduce plans for directions of future work and research in the area.

\subsection{Related Work}
\label{sec:related}

The foundation of our tool is based on the paper of Rikters et al. \cite{rikters2017visualizing}, who introduce visualization of NMT attention and use attention-based scoring of NMT as described by Rikters and Fishel \cite{riktersfishel2017}. While in general it can be useful to quickly find sentences with ``scrambled" attention alignments, it does have several flaws like considering completely untranslated sentences as good. This consistently misleads users when sorting data sets by confidence and looking for the highest scoring examples. Another shortcoming is the ability to only visualize a translation from one system at a time, making it slightly tricky to directly compare how multiple systems handle the same inputs.

In contrast, both iBLEU \cite{madnani2011ibleu} --- a web-based tool for visualizing BLEU \cite{Papineni2002} scores --- and MT-ComparEval \cite{klejch2015mt} --- which builds upon iBLEU by adding supplementary visualizations, scores and metrics --- can easily work with multiple MT outputs and even a set of human references. A downside for these tools is that the reference translation set is always mandatory and can't be left out. While it is useful to verify how the system performs in a controlled environment (when the expected result - reference translations - is known beforehand), more often than not the strangest abnormalities appear when using arbitrary data.

NMT frameworks like Nematus \cite{sennrich2017nematus}, Neural Monkey \cite{NeuralMonkey:2017} or OpenNMT \cite{2017opennmt} have some forms of visualization, but they mainly handle representation of the translation process instead of the translation results. For instance, OpenNMT has a separate repository for visualization tools\footnote{VisTools - https://github.com/OpenNMT/VisTools} that can generate visualizations of embeddings or beam search. Neural Monkey utilizes the built-in visualizations of TensorFlow \cite{abadi2016tensorflow} that can show the compute graph and multiple types of histograms from the training progress.

\section{Visualization Tool}
\label{sec:tool}

The basis of our visualization tool is described in full detail in the baseline paper \cite{rikters2017visualizing}. It requires source and translated sentences along with the corresponding attention alignments  from NMT systems as input files and can provide a visual overview in a command line environment (Linux Terminal or Windows Powershell) or a web browser of any modern device. It is published in a
GitHub repository\footnote{NMT Attention Alignment Visualizations: \url{https://github.com/M4t1ss/SoftAlignments}} 
and open-sourced with the MIT License. In the further subsections of the paper, we will outline only core components and focus more on highlighting improvements and differences.

In addition to Nematus, Neural Monkey and Marian\footnote{Marian: \url{https://github.com/marian-nmt/marian}}  \cite{junczys2018marian}, we have also added out-of-the-box support for working with attention alignments from OpenNMT and Sockeye\footnote{Sockeye: \url{https://github.com/awslabs/sockeye}} \cite{Sockeye:17} frameworks.

\subsection{Confidence Scores}
\label{sec:scores}

This section outlines how the confidence scores are calculated and outlines what is how the final score differs from the baseline.

\def\CP{{\rm CP}}
\def\CDP{{\rm CDP}}
\def\AP{{\rm AP}}
\def\OP{{\rm OP}}

The four main metrics that we use for scoring translations are:
\begin{itemize}
  \item \textbf{Coverage Deviation Penalty} (\CDP) penalizes attention deficiency and excessive attention per input token.
  \begin{equation}
	\CDP = -\frac{1}{L_{s}} \sum_j \log \left(1 + \Bigl(1 - \sum_i \alpha_{ji} \Bigr)^2 \right)
  \end{equation}

  \item \textbf{Absentmindedness Penalties} (\AP\textsubscript{out}, \textsubscript{in}) penalize output tokens that pay attention to too many input tokens, or input tokens that produce too many output tokens.
  
  \begin{equation}
	\AP_{out} = -\frac{1}{L_{s}} \sum_i \sum_j \alpha_{ji} \cdot \log \alpha_{ji}
  \end{equation}

  \begin{equation}
	\AP_{in} = -\frac{1}{L_{s}} \sum_j \sum_i \alpha_{ij} \cdot \log \alpha_{ij}
  \end{equation}

  \item \textbf{Overlap Penalty} (\OP) penalizes translations that copy large fractions from source sentences. A stronger penalty is allocated to longer sentences that copy large amounts from the source while shorter ones get more tolerance (e.g., the three-word English sentence ``Thanks Barack Obama." can be perfectly translated into ``Paldies Barack Obama." although 2/3 of words in the translation are the same in the source). 
%   An illustration of how the similarity penalty progresses depending on the sentence length and source-translation similarity is shown in Figure \ref{fig:similarity-penalty}. 

% Varbūt pārsaukt arī rakstā Similarity par Overlap (?!?)
% Labs piemērs līdzības sodam
% http://attention.lielakeda.lv/?directory=Compare%20Nematus%20-%20Neural%20Monkey%20200s%20En-%3ELv&s=42

  \begin{equation}
	\OP = (0.8+(L_{t}*0.01)) * (3-((1-S)*5)) * (0.7 + S) * tan(S)
  \end{equation}

% \begin{figure*}[t]
%   \resizebox {\textwidth} {!} {
%     \begin{tikzpicture}
%       \begin{axis}[
%           xlabel style={sloped},
%           ylabel style={sloped},
%           zlabel style={sloped},
%           xlabel={Overlap},
%           ylabel={Sentence length},
%           zlabel={Overlap penalty}
%       ]
%       \addplot3[
%           domain=0.3:1, 
%           y domain=0:300,
%           surf,
%       ]
%       {(0.8+(y*0.01)) * (3-((1-x)*5)) * (0.7 + x) * tan(deg(x))};
%       \end{axis}
%     \end{tikzpicture}
%   }
%   \caption{A plot of the similarity penalty.}
%   \label{fig:similarity-penalty}
% \end{figure*}

\item \textbf{Confidence} is the sum of the three main metrics -- \CDP{}, \AP{}\textsubscript{in} and \AP{}\textsubscript{out} and the similarity penalty, when the similarity between input and output sentences is high (similarity \textgreater \space 0.3) .
\end{itemize}
  \begin{equation}
    \scriptsize confidence = 
    \begin{cases}
        CDP + AP_{out} + AP_{in}, & \text{if } similarity<0.3 \\
        CDP + AP_{out} + AP_{in} - OP, & \text{otherwise}
    \end{cases}
  \end{equation}

In all of the metrics \(L_{s}\) is the length of the source sentence; \(L_{t}\) - length of the target sentence; S - similarity between the source sentence and the translation on the scale of 0 - 1; \(\alpha_{ji}\) - the attention weight between source token \textit{i} and translation token \textit{j}.

Changes have been introduced to the final confidence score by first calculating the similarity ratio between input and output sentences and then adding a further penalty only if the similarity is high enough. The similarity is calculated by finding the longest contiguous matching subsequence.

\begin{figure*}[t]
  \includegraphics[width=\linewidth]{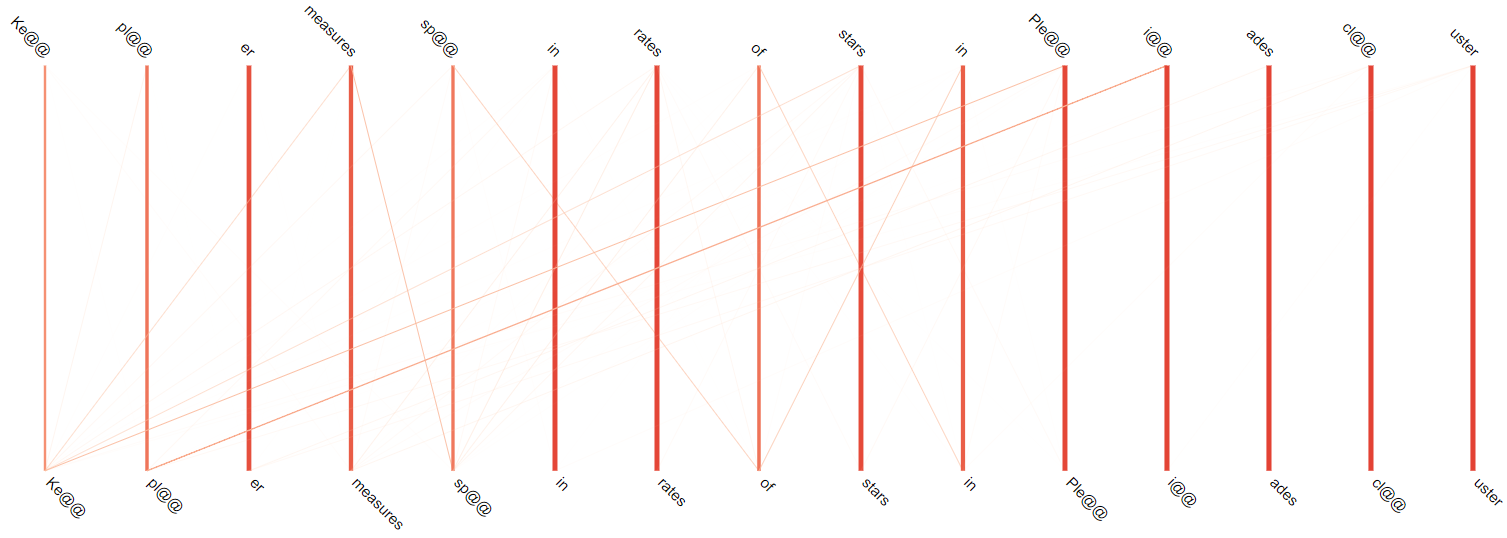}
  \begin{tabular}{lp{10.4cm}}
   \bf Source: & Kepler measures spin rates of stars in Pleiades cluster \\
   \bf Hypothesis: & Kepler measures spin rates of stars in Pleiades cluster \\
   \bf Reference: & Keplers izmēra zvaigžņu griešanās ātrumu Plejādes zvaigznājā. \\
  \end{tabular}
  \caption{An example of a translated sentence that exhibits a verbatim rendition of the input. CDP: 100.0\%; \(\AP_{out}\): 98.84\%; \(\AP_{in}\): 98.85\%; Baseline Confidence: \textbf{95.44\%}; Updated Confidence: \textbf{25.02\%}; }  
  \label{fig:less-confidence}
\end{figure*}

Since the baseline confidence score considered only the attention alignments when calculating the final value, examples like shown in Figure \ref{fig:less-confidence} received particularly high values due to consistent one-to-one attention alignments. The updated score takes care of this problem by penalizing hypothesis sentence that is overly similar to the input source. 

\subsection{Web Interface}
\label{sec:web}

The web interface is the primary point of interaction with the tool. Aside from browsing visualizations, ordering data sets by confidence scores and exporting visualizations as images, that are all clarified in the baseline paper, we introduce several significant changes to the system. The first one is a technical update on how data is served --- loading is performed asynchronously in the background and thereby eliminating long wait times to view the proceeding sentences in a large data set. 
The three major additions are:
  \begin{itemize}
    \item the addition of source-translation overlap percentage alongside the four base scores (Section \ref{sec:over});
    \item the ability to provide reference translations, if available, to display next to the hypothesis and calculate BLEU scores (Section \ref{sec:bleu});
    \item the ability to directly compare translations and alignments from two different NMT systems (Section \ref{sec:cmp}).
  \end{itemize}

% \begin{figure}[ht]
%   \includegraphics[width=\linewidth]{Figures/Document_level_scores.png}
%   \caption{Document-level scores for Confidence, CDP, APin and APout.}
%   \label{fig:document-scores}
% \end{figure}

% The second update is the addition of document-level scores for each one of the metrics. As shown in Figure \ref{fig:document-scores} - the scores are rendered in the same percentage values as sentence-level scores, with the only difference being that the values are averaged over the whole document.

\subsection{Overlap}
\label{sec:over}

\begin{figure}[ht]
  \includegraphics[width=\linewidth]{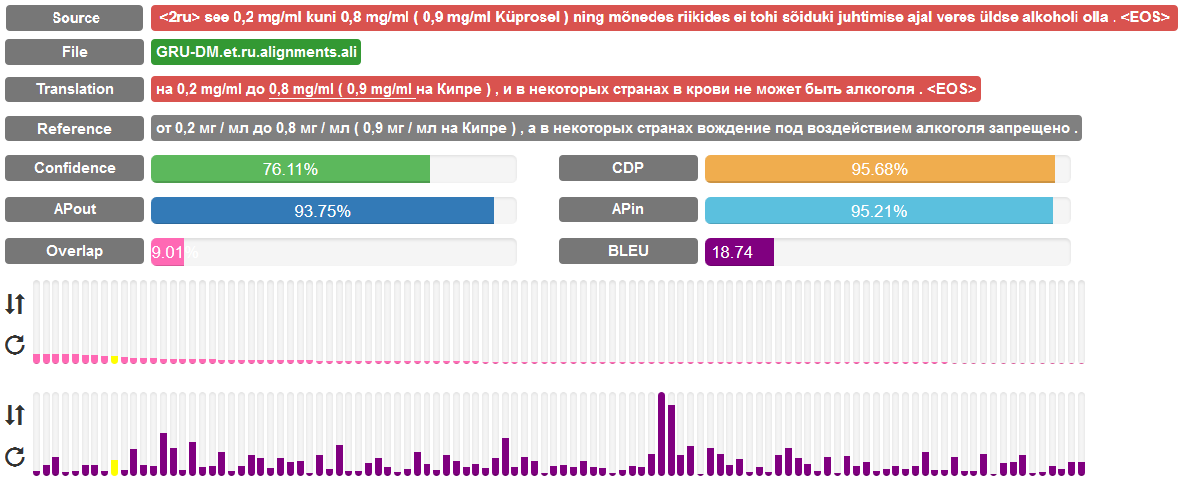}
  \begin{tabular}{lp{10.4cm}}
   \bf Source: & see 0,2 mg/ml kuni 0,8 mg/ml ( 0,9 mg/ml Küprosel ) ning mõnedes riikides ei tohi sõiduki juhtimise ajal veres üldse alkoholi olla.  \\
   \bf Hypothesis: & \foreignlanguage{russian}{на 0,2 mg/ml до 0,8 mg/ml ( 0,9 mg/ml на Кипре ) , и в некоторых странах в крови не может быть алкоголя.}  \\
   \bf Match: & 0,8 mg/ml ( 0,9 mg/ml \\
  \end{tabular}
  \caption{An example translation from Estonian into Russian, showing useful features for debugging translation outcomes - underlining of the longest matching substring between the source and translated sentences; sorting translations by overlap (pink bars) or BLEU score (purple bars); reference translation (gray background).}
  \label{fig:overlap-bleu}
\end{figure}

As mentioned in Section \ref{sec:scores}, the updated confidence score considers hypotheses translations that are long and have a significant overlap with the source sentence as a worse translations, while tolerating considerable overlap for shorter sentences. 
In addition to contributing to the final confidence score, the overlap ratio has been added as an individual score for sorting, navigating and comparing sentences from a data set as shown in Figure \ref{fig:overlap-bleu}.
The system also underlines the longest matching substring between the source and translation in cases where the overlap is high enough (over 10\%). An example is shown in Figure \ref{fig:overlap-bleu}, where the overlap ratio is 20.19\%. 

\subsection{References and BLEU}
\label{sec:bleu}

% One significant aspect that was missing from the baseline was any kind of comparison and scoring metric according to a set of given references. Although the authors stated that their tool does not require references and is supposed to be used as a reference-less method of MT evaluation, 
We believe that simply displaying the reference next to the hypothesis is helpful more often than not. Having provided references also allows to calculate BLEU scores for the translations, providing yet another dimension for sorting (Figure \ref{fig:overlap-bleu}). Unlike overlap, the BLEU scores do not influence the overall confidence scores.

% Both overlap and BLEU score calculation and output has also been added to the terminal interface of the tool (Figure \ref{fig:terminal}).

% \begin{figure}[ht]
%   \includegraphics[width=\linewidth]{Figures/terminal.png}
%   \caption{An example of the updated terminal interface output.}
%   \label{fig:terminal}
% \end{figure}

% \XXX{Add a screen-shot of the command-line output here}

\subsection{Comparing Translations}
\label{sec:cmp}

The final major addition to the tool is the option to directly compare two translations of the same source sentence. To perform the comparison, all source sentences for both input data sets must match, but the target sentences may differ in output token order as well as count. Comparisons may be performed between translations obtained from any two of the five currently supported NMT frameworks (Nematus, Neural Monkey, OpenNMT, Marian and Soceye) or even an arbitrary input file, as long as it's formatted according to the specification provided in the readme \footnote{Using other input formats - \url{https://github.com/M4t1ss/SoftAlignments\#how-to-get-alignment-files-from-nmt-systems}}.

\begin{figure*}[t]
  \includegraphics[width=\linewidth]{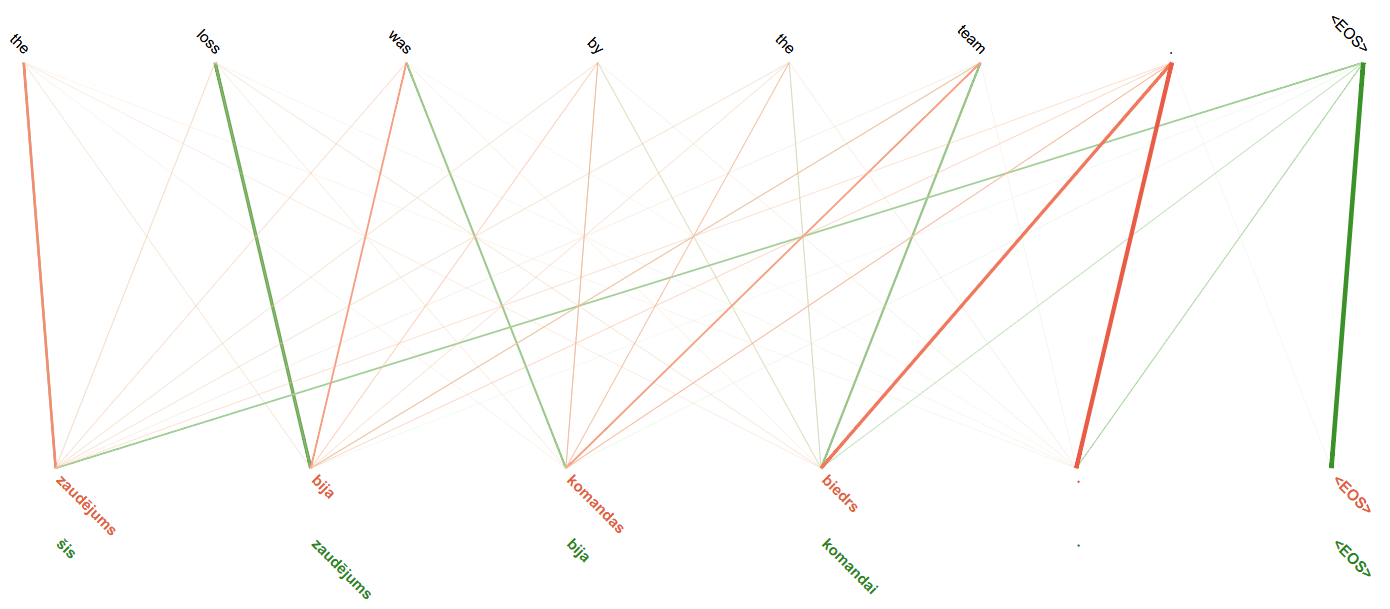}
  \begin{tabular}{lp{10.4cm}}
   \bf Source: & the loss was by the team. \\
   \bf Hypothesis 1: & zaudējums bija komandas biedrs. \\
   \bf Hypothesis 2: & šis zaudējums bija komandai. \\
   \bf Reference: & zaudē komanda. \\
  \end{tabular}
  \caption{A direct comparison of attention alignments for translating the same sentence with two different NMT systems.}
  \label{fig:nmt-comparison}
\end{figure*}

Figure \ref{fig:nmt-comparison} shows an example comparison of a sentence translated by two different NMT systems. On the top row is the source text and the bottom rows represent output from each individual NMT system color-coded to match the colors of the alignment lines. The second hypothesis (in green) exhibits stronger and more reliable output alignments to the content words while the first shows strong alignments coming from the stop sign. In this example neither hypothesis matches the reference, but since it is only two words long for a source sentence of triple the length, it can hint to an oversimplified translation by the translator (assuming English was the original) and does not mean that both hypotheses are completely wrong. In fact, the second hypothesis is a fairly decent representation of the source sentence.

\begin{figure*}[ht]
  \includegraphics[width=\linewidth]{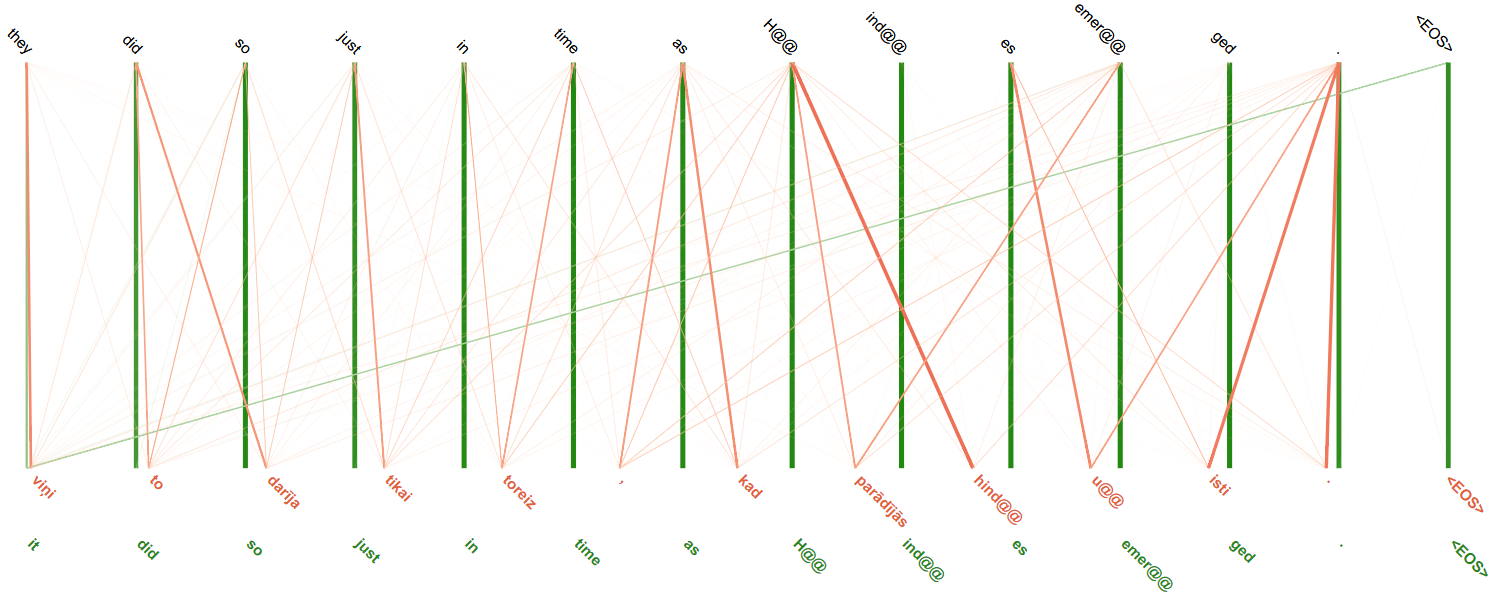}
  \begin{tabular}{lp{10.4cm}}
   \bf Source: & they did so just in time as Hindes emerged. \\
   \bf Hypothesis 1: & viņi to darīja tikai toreiz , kad parādījās hinduisti. \\
   \bf Hypothesis 2: & it did so just in time as Hindes emerged. \\
   \bf Reference: & viņiem tas izdevās pēdējā brīdī. \\
  \end{tabular}
  \caption{A comparison of lower and higher scoring hypotheses from two different NMT systems. 
  Scores for Hypothesis 1 (orange): Confidence \textbf{53.1\%}; Overlap \textbf{0.9\%}. Scores for Hypothesis 2 (green): Confidence \textbf{28.63\%}; Overlap \textbf{94.03\%}. }
  \label{fig:nmt-comparison-2}
\end{figure*}

Figure \ref{fig:nmt-comparison-2} illustrates another example with strong attention alignments and a high overlap ratio (94.03\%) between source and translated sentences from one system compared to a weak, but at least better translation from another system. The final confidence score for the second translation is strongly influenced by the high overlap, even though the sentence is not particularly long. In similar conditions, the confidence score of the second hypothesis calculated by the baseline system would be very close to 100\% due to its complete disregard for the actual words of the source and hypothesis sentences. 

\section{Recipes for Debugging}
\label{sec:debugging}

In this section we summarise several tips and tricks that may come in handy when using the tool to look for faulty translations of various kinds. Here we also list common causes associated with the problems. Some peculiarities to pay attention to may include:
  \begin{itemize}
    \item \textbf{Short sentences with a low confidence, CDP, \(\AP_{in}\) or \(\AP_{out}\)} 
    
    All of the metrics do not necessarily need to be low, but translations that exhibit at least one of them to be under 30\% are often worth looking into.
    \item \textbf{Long sentences with a high overlap}
    
    As stated before, for short, several words long sentences it may be completely normal to have an overlap of 50\% or more, but if it occurs in sentences that are 10 or more words long, it may indicate that the system has only partially translated the source or not translated anything at all. When completely untranslated sentences are found, it is worth checking the training data for any source-target sentence pairs that are equal. Removing them from the training data should help.
    \item \textbf{Sentences with a low BLEU score, but normal or even high confidence, CDP, \(\AP_{in}\) and \(\AP_{out}\)}
    
    The BLEU metric has its flaws and one of them is comparing each hypothesis to only one reference, while it is often possible to translate the same sentence in several different ways. In cases when the only low-scoring metric output by the tool is the BLEU score, it is often that the translation is perfectly good, but just different from the reference. Such sentences are often useful examples to show that lower BLEU scores of neural MT systems do not necessarily represent lower quality translations and are cheaper to find than performing full manual human evaluations.
  \end{itemize}

A separate recommendation specifically for comparing two translations is to look at the attention alignment lines and try to find ones with source tokens having strong alignments to different hypothesis tokens, while maintaining relatively similar confidence scores. Such translations are often synonyms.

\section{Conclusion}
\label{sec:conclusion}

In this paper, we described our conversion of a visualization tool into an instrument for debugging output form neural machine translation systems by improving the attention alignment scoring and confidence estimation of the baseline. The tool is intended to help researchers better understand how their systems perform by enabling to quickly locate better and worse translations in a arbitrary test sets.
Compared to other similar tools, ours relies on the confidence scores and does not require reference translations to facilitate this easier navigation, but it only benefits with additional features that are enabled when the references are provided. This allows to integrate it, for example, in an NMT system with a web interface, providing users with an explanation for the result of a specific translation.

In a future version of the system we may include other reference-based MT scoring metrics for more variety of scoring and sorting. Some examples of metrics may include chrF \cite{popovic2015chrf} or TER \cite{snover2006study}. Another idea for future work would be to list and order specific best, worst or interesting examples of translations. This could be done by considering the recipes from Section \ref{sec:debugging}. 

In addition to the reference-based metrics, there still are some reference-less approaches yet to be utilised. For instance, borrowing ideas from parallel corpora filtering \cite{pinnis-EtAl:2017:WMT} such as 1) source-hypothesis sentence length difference; 2) language identification for the hypothesis; 3) digit mismatch between the source and hypothesis; 4) foreign or corrupt symbol checking for the hypothesis.

\begin{figure*}[ht]
  \includegraphics[width=\linewidth]{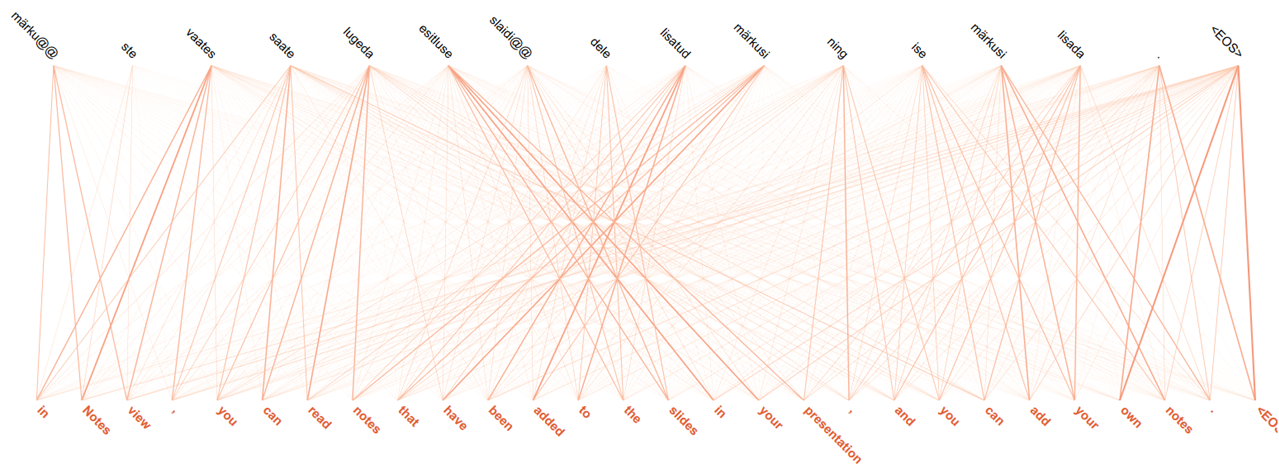}
  \caption{An example of attention alignments from a 15-layer encoder and 15-layer decoder convolutional neural machine translation system trained with FairSeq.}
  \label{fig:multi-layer}
\end{figure*}

Another ongoing challenge is to find a way of better representing attention alignments generated by multi-layer neural networks. While in recurrent neural network NMT systems this is rarely a problem, more modern approaches like convolution neural networks \cite{gehring2017convolutional} and transformer neural networks \cite{VaswaniSPUJGKP17} require training of deeper models to achieve competitive quality translation results. This, however, results in each layer paying attention only to a subset of the input sentence. Even when all attentions are summed up, the result looks like every source token is connected to every hypothesis token as can be seen in Figure \ref{fig:multi-layer}.

% In the full paper we plan to perform a manual human evaluation on the updated confidence score to measure improvements in correlation with human judgments. After the human evaluation we will analyse some of the highest and lowest correlating examples in more detail. 

\section{Acknowledgments}
\label{sec:acknowledgments}

The research has been supported by the European Regional  Development  Fund  within  the  research project ”Neural Network Modelling for Inflected Natural Languages” No. 1.1.1.1/16/A/215.

\FloatBarrier

% \section{Bibliographical References}
% \label{sec:ref}

\bibliographystyle{lncs/splncs04}
\bibliography{bibliography}

\end{document}